\documentclass[twoside,11pt]{article}
\usepackage{jmlr2e}
\usepackage{amsmath,amsfonts,amssymb}

\def\vec#1{\ensuremath{\mathchoice
                     {\mbox{\boldmath$\displaystyle\mathbf{#1}$}}
                     {\mbox{\boldmath$\textstyle\mathbf{#1}$}}
                     {\mbox{\boldmath$\scriptstyle\mathbf{#1}$}}
                     {\mbox{\boldmath$\scriptscriptstyle\mathbf{#1}$}}}}

\graphicspath{{pic/}}

\ShortHeadings{Quantifying and Visualizing Attribute Interactions}{Jakulin
\& Bratko } \firstpageno{1}

\def\aive{a\"\i{}ve }

\begin{document}

\title{Quantifying and Visualizing Attribute Interactions:\\ An Approach Based on Entropy}

\author{\name Aleks Jakulin \email jakulin@acm.org \\
       \name Ivan Bratko \email ivan.bratko@fri.uni-lj.si \\
       \addr Faculty of Computer and Information Science\\ University of Ljubljana,
       Tr\v{z}a\v{s}ka cesta 25\\ SI-1001 Ljubljana, Slovenia}

\editor{}

\maketitle

\begin{abstract}
Interactions are patterns between several attributes in data that cannot
be inferred from any subset of these attributes. While mutual information
is a well-established approach to evaluating the interactions between two
attributes, we surveyed its generalizations as to quantify interactions
between several attributes. We have chosen McGill's interaction
information, which has been independently rediscovered a number of times
under various names in various disciplines, because of its many
intuitively appealing properties. We apply interaction information to
visually present the most important interactions of the data.
Visualization of interactions has provided insight into the structure of
data on a number of domains, identifying redundant attributes and
opportunities for constructing new features, discovering unexpected
regularities in data, and have helped during construction of predictive
models; we illustrate the methods on numerous examples. A machine learning
method that disregards interactions may get caught in two traps: myopia is
caused by learning algorithms assuming independence in spite of
interactions, whereas fragmentation arises from assuming an interaction in
spite of independence.
\end{abstract}


\section{Introduction}
\label{sec:introduction}

One of the basic notions in probability is the concept of
\emph{independence}. Binary events $a$ and $b$ are independent if and only
if $P(a,b)=P(a)P(b)$. Independence also implies that $a$ is irrelevant to
$b$. However, independence is not a stable relation: $a$ may become
dependent with $b$ if we observe another event $c$. For example, define
$c$ to happen when $a$ and $b$ take place together ($a \wedge b$), or do
not take place together ($\neg a \wedge \neg b$). Even if $a$ and $b$ are
independent and random, they become dependent in the context of $c$.
Alternatively, $a$ may become independent of $b$ in the context of $c$,
even if they were dependent before: imagine that $a$ and $b$ are two
independently sampled uncertain measurements of $c$. Without $c$, the
measurements are similar, hence their dependence. With the knowledge of
$c$, however, the similarity between $a$ and $b$ disappears. It is hence
difficult to systematically investigate dependencies.

\emph{Conditional independence} was proposed as a solution to the above
problem of flickering dependencies. Events $a$ and $b$ are conditionally
independent with respect to event $c$ if and only if they are independent
in the context of every outcome of $c$. In probability theory, this can be
expressed as $P(a|b,c)=P(a|c)$. Hence, independence cannot be claimed
unless it has been verified in the context of all other events. This helps
us simplify the model of the environment.

In this text, we endorse a different type of regularities,
\emph{interactions}, which subsumes conditional independence. An
interaction is a regularity, a pattern, a dependence present only in the
whole set of events, but not in any subset. When there is such a
regularity, we say that the attributes of the set interact. For example, a
2-way interaction between two events indicates that the joint probability
distribution cannot be described with the assumption of mutual
independence between events. A 3-way interaction between three events is
equivalent to the inability to describe the joint probability distribution
with any marginalization, it is hence necessary to model it directly.
Interactions are local, meaning that they are only defined in the context
of events they relate to. Interactions are stable, because introduction of
newly observed events cannot change the interactions that already exist
among the events. An interaction is unambiguous, meaning that there is
only one way of describing it. Interactions are symmetric and undirected,
so directionality no longer needs to be explained by, e.g., causality.

The contributions of each successive section of this paper can be
summarized as follows:
\begin{itemize}
\item A survey of information-based measures of interaction among
attributes.
\item Several novel diagrams for visualization of interactions
in the data.
\item Investigation of relevance of interactions in the
context of supervised machine learning.
\end{itemize}

\section{Quantifying Interactions with Entropy}

In this section, we will revise the basic concepts of information theory
in order to derive \emph{interaction information}, our proposal for
quantifying higher-order dependencies in data. Interaction information
captures and quantifies the earlier intuitive view of probabilistic
interactions. The interdisciplinary review of related work at the end of
the section shows that virtually the same formulae have emerged
independently a number of times in various disciplines, ranging from
physics to psychology, adding weight to the worth of the idea.

\subsection{Attributes, Probabilities and Entropy}

While we used the terminology of events earlier, we will now migrate to
common machine learning terminology. An \emph{attribute} $A$ will be
considered to be a collection of independent but mutually exclusive
events, or attribute values, $\{a_1,a_2,a_3,\ldots,a_n\}$. We will
consider $a$ as an example of an event from $A$'s alphabet, or $A$'s
value. Variates, random variables, communication sources and classifiers
can all be considered to be types of attributes. In machine learning, an
\emph{instance} corresponds to another event, which is described as a set
of attributes' events. For example, an instance is ``Dancing in cold
weather, in rain and in wind.'' and such instances are described with four
attributes, $A$ with the alphabet or range
$\Re_A=\{\text{dancing},\neg\text{dancing}\}$, $B: \Re_B =
\{\text{cold},\text{cool},\text{warm},\text{hot}\}$, $C:
\{\text{rain},\text{cloud},\text{clear}\}$, and $D:
\{\text{calm},\text{breeze},\text{wind}\}$. If our task is deciding
whether to dance or not to dance, attribute $A$ is the \emph{label}.

An instance is a synchronous observation of the attributes, and we can
describe the relationships between the attributes, assuming that the
instances are permutable, with a \emph{joint probability distribution}
$P(A,B,C)$. \emph{Marginal probability distributions} are projections of
the joint where we disregard but a subset of attributes, for example:
$$P(A) = P(A,\cdot,\cdot) = \sum_{b \in \Re_B}\sum_{c \in \Re_C}P(A,b,c).$$
% repeat the definition from Sec.1, detailing it
The essence of learning is simplification of the joint probability
distributions, achieved by exploiting certain regularities. A very useful
discovery is that $A$ and $B$ are independent, meaning that $P(A,B)$ can
be approximated with $P(A)P(B)$. If so, we say that $A$ and $B$ do not
2-interact, or that there is no 2-way interaction between $A$ and $B$.
Unfortunately, attribute $C$ may affect the relationship between $A$ and
$B$ in a number of ways. Controlling for the value of $C$, $A$ and $B$ may
prove to be dependent even if they were previously independent. Or, $A$
and $B$ may actually be independent when controlling for $C$, but
dependent otherwise.

If the introduction of the third attribute $C$ affects the dependence
between $A$ and $B$, we say that $A$, $B$ and $C$ 3-interact, meaning that
we cannot decipher their relationship without considering all of them at
once. An appearance of a dependence is an example of a \emph{positive
interaction}: positive interactions imply that the introduction of the new
attribute increased the amount of dependence. A disappearance of a
dependence is a kind of a \emph{negative interaction}: negative
interactions imply that the introduction of the new attribute decreased
the amount of dependence. If $C$ does not affect the dependence between
$A$ and $B$, we say that there is no 3-interaction.

There are plenty of real-world examples of interactions. Negative
interactions imply redundance. For example, weather attributes rain and
lightning are dependent, because they occur together. But the attribute
storm interacts negatively with them, since it reduces their dependence.
Storm explains a part of their dependence. Should we wonder whether there
is lightning, the information that there is rain would contribute no
information if we already knew that there is a storm.

Positive interactions imply synergy instead. For example, employment of a
person and criminal behavior are not particularly dependent attributes
(most unemployed people are not criminals, and many criminals are
employed), but adding the knowledge of whether the person has a new sports
car suddenly makes these two attributes dependent: it is a lot more
frequent that an unemployed person has a new sports car if he is involved
in criminal behavior; the opposite is also true: it is somewhat unlikely
that an unemployed person will have a new sports car if he is not involved
in criminal behavior.

In real life, it is quite rare to have perfectly positive or perfectly
negative interactions. Instead, we would like to quantify the magnitude
and the type of an interaction. For this, we will employ \emph{entropy} as
a measure of uncertainty. A measure of mutual dependence can be
constructed from an uncertainty measure, defining dependence as the amount
of shared uncertainty.

Let us assume an attribute, $A$. Shannon's entropy measured in bits is a
measure of unpredictability of an attribute~\citep{Shannon48}:
\begin{equation}
    H(A)\triangleq  -\sum_{a \in \Re_A}P(a)\log_2{P(a)}
\end{equation}
By definition, $0\log_2 0=0$. The higher the entropy, the less reliable
are our predictions about $A$. We can understand $H(A)$ as the amount of
uncertainty about $A$, as estimated from its probability distribution.
Although this definition is appropriate only for discrete sources, or
discretized continuous sources, \citet{Shannon48} also presents a direct
definition for continuous ones.

\subsection{Entropy Calculus for Two Attributes}

Let us now introduce a new attribute, $B$. We have observed the joint
probability distribution, $P(A,B)$. We are interested in predicting $A$
with the knowledge of $B$. At each value of $B$, we observe the
probability distribution of $A$, and this is expressed as a conditional
probability distribution, $P(A|B)$. Conditional entropy, $H(A|B)$,
quantifies the remaining uncertainty about $A$ with the knowledge of $B$:
\begin{equation}
    H(A|B)\triangleq -\sum_{a \in \Re_A, b \in \Re_B}{P(a,b)\log_2{P(a|b)}}=H(A,B)-H(B)
\end{equation}
We quantify the 2-way interaction between two attributes with \emph{mutual
information}:
\begin{equation}
    \begin{split}
    I(A;B)\triangleq \sum_{a \in \Re_A, b \in \Re_B}{P(a,b)\log_2\frac{P(a,b)}{P(a)P(b)}}
    = H(A)+H(B)-H(A,B)\\ = H(A)-H(A|B) = I(B;A) = H(B)-H(B|A)
    \end{split}
\end{equation}
In essence, $I(A;B)$ is a measure of correlation between attributes, which
is always zero or positive. It is zero if and only if the two attributes
are independent, when $P(A,B)=P(A)P(B)$. Observe that the mutual
information between attributes is the \emph{average} mutual information
between events in attributes' alphabets. If $A$ is an attribute and $L$ is
the label attribute, $I(A;L)$ measures the amount of information provided
by $A$ about $L$: in this context it is often called \emph{information
gain}.

\begin{figure}
\begin{center}
\vspace{-15mm}
\includegraphics[width=5.5cm]{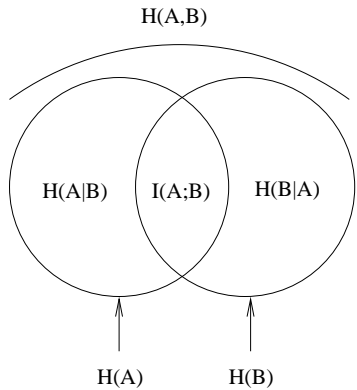}
\vspace{-17mm}
\end{center}
\caption{A graphical illustration of the relationships between
information-theoretic measures of the joint distribution of attributes $A$
and $B$. The surface area of a section corresponds to the labelled
quantity. This illustration is inspired by \cite{CoverThomas}.
\label{fig:venn2}}
\end{figure}

A 2-way interaction helps reduce our uncertainty about either of the two
attributes with the knowledge of the other one. We can calculate the
amount of uncertainty remaining about the value of $A$ after introducing
knowledge about the value of $B$. This remaining uncertainty is $H(A|B)$,
and we can obtain it using mutual information, $H(A|B) = H(A)-I(A;B).$
Sometimes it is worth expressing it as a percentage, something that we
will refer to as \emph{relative} mutual information. For example, after
introducing attribute $B$, we have $100\%\cdot H(A|B)/H(A)$ percent of
uncertainty about $A$ remaining. For two attributes, the above notions are
illustrated in Fig.~\ref{fig:venn2}.

\subsection{Entropy Calculus for Three Attributes}

Let us now introduce the third attribute, $C$. We could wonder how much
uncertainty about $A$ remains after having obtained the knowledge of $B$
and $C$: $H(A|BC) = H(ABC)-H(BC).$ We might also be interested in seeing
how $C$ affects the interaction between $A$ and $B$. This notion is
captured with \emph{conditional mutual information}:
\begin{equation}
    \begin{split}
    I(A;B|C)\triangleq \sum_{a, b, c}{P(a,b,c)\log_2\frac{P(a,b|c)}{P(a|c)P(b|c)}} = H(A|C) + H(B|C) - H(AB|C) \\
    = H(A|C)-H(A|B,C)  = H(AC) + H(BC) - H(C) - H(ABC).
    \label{eq:cmi3}
    \end{split}
\end{equation}
Conditional mutual information is always positive or zero; when it is
zero, it means that $A$ and $B$ are unrelated given the knowledge of $C$,
or that $C$ completely explains the association between $A$ and $B$. From
this, it is sometimes inferred that $A$ and $B$ are both consequences of
$C$. If $A$ and $B$ are conditionally independent, we can apply the n\aive
Bayesian classifier for predicting $C$ on the basis of $A$ and $B$ with no
remorse. Conditional mutual information is a frequently used heuristic for
constructing Bayesian networks \citep{Cheng02}.

Conditional mutual information $I(A;B|C)$ describes the relationship
between $A$ and $B$ in the context of $C$, but we do not know the amount
of influence resulting from the introduction of $C$. This is achieved by
the measure of the intersection of all three attributes, or
\emph{interaction information}~\citep{McGill54} or \emph{McGill's multiple
mutual information}~\citep{han80}:
\begin{equation}
    \begin{split}
    I(A;B;C)\triangleq I(A;B|C)-I(A;B)=I(A,B;C)-I(A;C)-I(B;C)\\
    =H(AB)+H(BC)+H(AC)-H(A)-H(B)-H(C)-H(ABC).
    \label{eq:ig3}
    \end{split}
\end{equation}
Interaction information among attributes can be understood as the amount
of information that is common to all the attributes, but not present in
any subset. Like mutual information, interaction information is symmetric,
meaning that $I(A;B;C)=I(A;C;B)=I(C;B;A)=\ldots$. Since interaction
information may be negative, we will often refer to the absolute value of
interaction information as \emph{interaction magnitude}. Again, be warned
that interaction information among attributes is the average interaction
information among the corresponding events.

The concept of \emph{total correlation}~\citep{Watanabe60} describes the
total amount of dependence among the attributes:
\begin{equation}
    \begin{split}
    C(A,B,C) \triangleq H(A)+H(B)+H(C)-H(ABC) \\
    = I(A;B)+I(B;C)+I(A;C)+I(A;B;C).
    \end{split}
\end{equation}
It is always positive, or zero if and only if all the attributes are
independent, $P(A,B,C)=P(A)P(B)P(C)$. However, it will not be zero even if
only a pair of attributes are dependent. For example, if
$P(A,B,C)=P(A,B)P(C)$, the total correlation will be non-zero, but only
$A$ and $B$ are dependent. Hence, it is not justified to claim an
interaction among all three attributes. For such a situation, interaction
information will be zero, because $I(A;B|C)=I(A;B)$.

\subsubsection{Positive and negative interactions}

Interaction information can either be positive or negative. Perhaps the
best way of illustrating the difference is through the equivalence
$I(A;B;C)=I(A,B;C)-I(A;C)-I(B;C)$: Assume that we are uncertain about the
value of $C$, but we have information about $A$ and $B$. Knowledge of $A$
alone eliminates $I(A;C)$ bits of uncertainty from $C$. Knowledge of $B$
alone eliminates $I(B;C)$ bits of uncertainty from $C$. However, the joint
knowledge of $A$ and $B$ eliminates $I(A,B;C)$ bits of uncertainty. Hence,
if interaction information is positive, we benefit from a synergy. A
well-known example of such synergy is the exclusive or: $C=A+B \pmod 2$.
If interaction information is negative, we suffer diminishing returns by
several attributes providing overlapping, redundant information. Another
interpretation, offered by \citet{McGill54}, is as follows: Interaction
information is the amount of information gained (or lost) in transmission
by controlling one attribute when the other attributes are already known.

\subsubsection{Interactions and Supervised Learning}
\label{sec:supervised-int}

The objective of unsupervised machine learning is the construction of a
model which helps predict the value of any attribute with partial
knowledge of other attribute values. For attributes $A,B,C$, unsupervised
models approximate the joint probability distribution $P(A,B,C)$ with a
joint probability distribution function $\hat{P}(A,B,C)$. On the other
hand, the objective of supervised learning is to predict the distinguished
label attribute with the (partial) knowledge of other attributes.
Supervised models attempt to describe the conditional probability
distribution, distinguishing the label $C$ from ordinary attributes $A$
and $B$. The conditional probability distribution can be modelled with
informative models $\hat{P}(A,B|C)$, or with discriminative models
$\hat{P}(C|A,B)$ \citep{Rubinstein97}. For example, the n\aive Bayesian
classifier is an informative model, while logistic regression is a
discriminative model.

In supervised learning we are primarily interested in interactions that
involve the label: interactions between non-label attributes alone are
rarely investigated. In fact, only interactions involving the label can
provide any information about it. If there is no interaction with the
label, there is no information about the label. As an example, we
formulate the n\aive Bayesian classifier as an approximation to the Bayes
rule, introducing the assumption that $A$ and $B$ are independent given
$C$, $P(A,B|C) \approx P(A|C)P(B|C)$:
\begin{equation}
    P(C|A,B) = P(C)\frac{P(A,B|C)}{P(A,B)} \approx
    P(C)\frac{P(A|C)P(B|C)}{P(A,B)}
    \label{eq:n-bayes}
\end{equation}
The conditional independence assumption is that there does not exist any
value of $C$, in the context of which $A$ and $B$ would 2-interact. We
will refer to this type of interactions as \emph{informative conditional
interactions}, the interactions in informative probability distributions.
For a label $C$ and attributes $X$ and $Y$, the 2-way informative
conditional interaction information is $I(X;Y|C)$, the familiar
conditional mutual information. It can be seen as the expected 2-way
informative conditional interaction information between $A$ and $B$ over
the values of $C$.

The relationship between the ordinary and the informative conditional
interactions is easily seen from the definition of 3-way interaction
information in \eqref{eq:ig3}: it is the difference between the two kinds
of 2-way interaction information. It is quite easy to see that when
$I(A;B|C)=0$, the 3-way interaction information can only be nonnegative.
When the 3-way interaction information is positive, the 2-way conditional
interaction information must also be positive. However, when the 3-way
interaction information is zero or negative, no specific conclusions can
be made about the 2-way informative conditional interaction.

\subsubsection{Limitations of the Bayesian network representation of conditional independence relations}

Given three attributes, if any pair of attributes is conditionally
independent given the third, e.g., $I(A;B|C)=0$, the interaction
information among the three cannot be positive. Such a situation can be
perfectly represented with a Bayesian network~\citep{Pearl88}: $A
\leftarrow C \rightarrow B$. If a pair of attributes are mutually
independent, for example $I(A;B)=0$, the interaction information, if it
exists, can only be positive. Such an interaction can also be perfectly
described with a Bayesian network, $A \rightarrow C \leftarrow B$.

Unfortunately, there are situations that can be described by several
Bayesian networks, none of which is able to describe the structure of
interactions. Assume the exclusive or problem, where $A = B + C \pmod 2$
with binary attributes. The mutual information between any pair of
attributes is zero, yet the attributes are in a 3-way interaction. We can
formally describe this with three consistent Bayesian networks: $A
\rightarrow C \leftarrow B,$ $A \rightarrow B \leftarrow C$ and $B
\rightarrow A \leftarrow C$, but no network emphasizes the fact that there
are no 2-way interactions. The interaction information in this case,
however, is strictly positive.

Another example is the case of triplicated attributes, $A = B = C$. In
this case, for all combinations of attributes, the conditional mutual
information is zero. On the other hand, every pair of attributes is
deterministically 2-way interacting. Again, the fact cannot be seen from
any of the Bayesian networks consistent with the data: $A \leftarrow C
\rightarrow B$, $A \leftarrow B \rightarrow C$ and $B \leftarrow A
\rightarrow C$. The interaction information in this case is strictly
negative.

These two were extreme examples, but having zero conditional mutual
information is too an extreme example. Conditional or mutual information
is rarely zero, yet some are larger than others. The larger they are the
more likely it is that the dependencies are not coincidental, and the more
information we gain by not ignoring them. We quantify interactions for
this precise reason, and our visualization methods emphasize the
quantified interaction magnitude.

\subsection{Quantifying $n$-Way Interactions}

In this section, we will generalize the above concepts to interactions
involving an arbitrary number of attributes. Assume a set of attributes
${\cal A}=\{X_1,X_2,\ldots,X_n\}$. Each attribute $X \in {\cal A}$ has an
alphabet $\Re_X=\{x_1,x_2,\ldots,x_p\}$. If we consider the whole set of
attributes ${\cal A}$ as a multivariate or a vector of attributes, we have
a joint probability distribution, $P(\vec{a})$. $\Re_{\vec{A}}$ is the
Cartesian product of individual attributes' alphabets, $\Re_{\vec{A}} =
\Re_{X_1} \times \Re_{X_2} \times \cdots \times \Re_{X_n}$, and
$\vec{a}\in \Re_{\vec{A}}$.  We can then define a marginal probability
distribution for a subset of attributes ${\cal S} \subseteq {\cal A}$,
where ${\cal S}=\{X_{i(1)},X_{i(2)},\ldots,X_{i(k)}\}$:
\begin{equation}
    P(\vec{s}) \triangleq
        \sum_{\substack{\vec{a}\in \Re_{\vec{A}},\\ \vec{s}_j = \vec{a}_{i(j)},\\ j = 1,2,\ldots,k}}
            P(\vec{a}).
\end{equation}
Next, we can define the entropy for a subset of attributes:
\begin{equation}
    H({\cal S}) \triangleq -\sum_{\vec{v} \in \vec{\cal S}}P(\vec{v})\log_2{P(\vec{v})}
\end{equation}
We define \emph{$k$-way interaction information} by generalizing from
formulae in \citep{McGill54} for $k=3,4$ to an arbitrary $k$:
\begin{equation}
I({\cal S}) \triangleq -\sum_{{\cal T}\subseteq {\cal S}}{(-1)^{|{\cal S}|
- |{\cal T}|}H({\cal T})} = I({\cal S}\setminus X|X) - I({\cal S}\setminus
X),\, X\in {\cal S},
\end{equation}
$k$-way multiple mutual information is closely related to the
lattice-theoretic derivation of multiple mutual information \citep{han80},
$\Delta h({\cal S})=-I({\cal S})$, and to the set-theoretic derivation of
multiple mutual information \citep{yeung91} and co-information
\citep{Bell02} as $I'({\cal S})=(-1)^{|{\cal S}|}I({\cal S})$.

Finally, we define \emph{$k$-way total correlation} as
\citep{Watanabe60,han80}:
\begin{equation}
C({\cal S}) \triangleq \sum_{X \in {\cal S}}H(X) - H({\cal S}) =
\sum_{{\cal T}\subseteq {\cal S}, |{\cal T}|\geq 2}{I({\cal T})}.
\end{equation}
We can see that it is possible to arrive at an estimate of total
correlation by summing all the interaction information existing in the
model. Interaction information can hence be seen as a decomposition of a
$k$-way dependence into a sum of $l, \, l\leq k$ dependencies.

\subsection{Related Work}

Although the idea of mutual information has been formulated (as `rate of
transmission') already by \citet{Shannon48}, the seminal work on
higher-order interaction information was done by~\citet{McGill54}, with
application to the analysis of contingency table data collected in
psychometric experiments, trying to identify multi-way dependencies
between a number of variables. The analogy between variables and
information theory was derived from viewing each variable as an
information source. These concepts have also appeared in biology at about
the same time, as \citet{Quastler53} gave the same definition of
interaction information as McGill, but with a different sign. The concept
of interaction information was also discussed in early textbooks on
information theory~\citep[e.g.][]{Fano61}. A formally rigorous study of
interaction information was a series of papers by Han, the best starting
point to which is \citep{han80}. A further discussion of mathematical
properties of positive versus negative interactions appeared
in~\citep{Tsujishita95}. \citet{Bell02} discussed the concept of
co-information, closely related to the Yeung's notion of multiple mutual
information, and suggested its usefulness in the context of dependent
component analysis.

In physics, \citet{Cerf97} associated positive interaction information of
three variables (referred to as ternary mutual information) with the
non-separability of a system in quantum physics. \citet{Matsuda00} applied
interaction information (referred to as higher-order mutual information)
and the positive/negative interaction dichotomy to the study of many-body
correlation effects in physics, and pointed out an analogy between
interaction information and Kirkwood superposition approximation. In
ecology, \citet{Orloci03} referred to interaction information as `the
mutual portion of total diversity' and denoted it as $I(ABC)$. In the
field of neuroscience, \citet{Brenner00} noted the utility of interaction
information for three attributes, which they referred to as synergy. They
used interaction information for observing relationships between neurons.
\citet{Gat99} referred to positive interactions as synergy, while and to
negative interactions as redundance. The concept of interactions also
appeared in cooperative game theory with applications in economics and
law. The issue is observation of utility of cooperation to different
players, for example, a coalition is an interaction between players which
might either be of negative or positive value for them.
\citet{InteractAxiom} formulated the Banzhaf interaction index, which
proves to be a generalization of interaction information, if negative
entropy is understood as game-theoretic value, attributes as players, and
all other players are disregarded while evaluating a coalition of a subset
of them \citep{JakulinMAG}. \citet{gd_measure} applied these notions to
rough set analysis.

The topic of interactions was a topic of extensive investigation in
statistics, and our review will be an extremely limited one.
\citet{Darroch74} surveyed two definitions of interactions, the
multiplicative, which was introduced by \citet{Bartlett35} and generalized
by \citet{Roy56}, and the additive definition due to \citet{Lancaster69}.
\citet{Darroch74} preferred the multiplicative definition, and described a
`partition' of interaction which is equivalent to the entropy-based
approach described in this paper. Other types of partitioning were
discussed by \citet{Lancaster69}, and more recently by \citet{Amari01} in
the context of information geometry.

\citet{Watanabe60} was one of the first to discuss total correlation in
detail, even if the same concept had been described (but not named)
previously by~\citet{McGill54}. \citet{Palus94} refers to it as
redundancy. \citet{studeny98multiinformation} investigated the properties
of conditional mutual information as applied to conditional independence
models. They discussed total correlation, generalized it, and named it
multiinformation. Multiinformation was used to show that conditional
independence models have no finite axiomatic characterization. More
recently \citet{Wennekers03} have referred to total correlation as
stochastic interaction, and \citet{Sporns00} as integration.
\citet{Chechik01} investigated the similarity between total correlation
and interaction information. \citet{vedral1} has compared total
correlation with interaction information in the context of quantum
information theory. Total correlation is sometimes even referred to as
multi-variate mutual information \citep{Boes99}.
% \citet{Meo02} has recently proposed a different generalization of mutual information,
% called the maximum independence estimate, derived by generalizing from
% $I(A;B)=H(A)-H(A|B)$.

%\citep{pluim:2003-732} discuss applications of mutual information in
%medical image registration.

\subsubsection{The relationship between set theory and entropy\label{sect:set-entropy}}
Interaction information is similar to the notion of intersection of three
sets. It has been long known that these computations resemble the
inclusion-exclusion properties of set theory \cite{yeung91}. We can view
mutual information $(;)$ as a set-theoretic intersection $(\cap)$, joint
entropy $(,)$ as a set-theoretic union $(\cup)$ and conditioning $(|)$ as
a set difference $(-)$. The notion of entropy or information corresponds
to $\mu$, a signed measure of a set, which is a set-additive function.
Yeung defines $\mu$ to be an I-measure, where $\mu^\ast$ of a set is equal
the entropy of the corresponding probability distribution, for example
$\mu^\ast(\tilde{X})=H(X)$. Yeung refers to diagrammatic representations
of a set of attributes as \emph{information diagrams}, similar to Venn
diagrams. Some think that using these diagrams for more than two
information sources is misleading~\citep{MacKay03}, for example because
the set measures can be negative, because there is no clear concept of
what the elements of the sets are, and because it is not always possible
to keep the surface areas proportional to the actual uncertainty.

Through the principle of inclusion-exclusion and understanding that
multiple mutual information is equivalent to an intersection of sets, it
is possible to arrive to a slightly different formulation of interaction
information \citep[e.g.][]{yeung91}. This formulation is the most frequent
in recent literature, but it has a counter-intuitive semantics, as
illustrated by \citet{Bell02}: $n$-parity, a special case of which is the
XOR problem for $n=2$, is an example of a purely $n$-way dependence. It
has a positive co-information when $n$ is even, and a negative
co-information when $n$ is odd. For that reason we decided to adopt the
original definition of \citep{McGill54}.

\subsubsection{Testing the significance of dependencies and interactions}
\citet{McGill54} discussed that expressions involving entropy are closely
associated with likelihood ratio; total correlation and conditional mutual
information follow a $\chi^2$ distribution for large samples.
\citet{han80} also discussed asymptotic properties of interaction
information. Entropy is a general statistic, useful for ascertaining
significance of various forms of dependence or independence among
attributes.

\subsubsection{Total and partial correlation}
Entropy is just a specific approach to quantification of variance, and
the inferences we make have much in common with those based upon partial
correlation. Our visualization methods would be equally suitable for
presenting partial correlation among continuous variables.

Partial correlation quantifies the amount of variance in the dependent
variable explained by only one independent variable, controlling for a
set of other independent variables \citep{Yule07}. If correlation can be
seen as a measure of dependence between a dependent variable and an
independent variable, multiple correlation as a measure of dependence
between a dependent variable and a number of independent variables, we
can understand partial correlation as a quantification of conditional
dependence for continuous variables.

Venn diagrams, such as the one in Fig.~\ref{fig:venn2} are also an item of
interest in statistics, where multiple correlation or the $R^2$ or
$\eta^2$ statistics represents the total amount of variance in the
dependent variable explained in the sample data by the independent
variables. With this index, it is too possible to render Venn diagrams
representing shared variance, even if these indices do not have properties
as appealing as entropy.

%%%%%%%%%%%%%%%%%%%%%%%%%%%%%%%%%%%%%%%%%%%%%%%%%%%%%%%%%%%%%%%%%%%%%%%%%%%%%%%%%%%%

\section{Visualizing Interactions}
\label{sec:visualizing}

Interactions among attributes are often very interesting for a human
analyst~\citep{Freitas01}. We will propose a number of novel diagrams in
this section to visually present interactions in data, providing examples
of quantification of interactions. Visualization methods attempt to
present the most important interactions that exist in the data. Entropy
and interaction information yield easily to graphical presentation, as
they are both measured in bits. Nonetheless, all the visualization methods
could be easily used with the measures of interaction magnitude. The
optimal type of visualization method depends on the type of learning it
supports. In unsupervised learning, we are interested in general
relationships between attributes. In supervised learning, we are
particularly interested in the relevance of individual attributes to
predictions about the label.

In our analysis, we have used the `census/adult', `mushroom', `pima',
`zoo', `Reuters-21578' and `German credit' data sets from the UCI
repository~\citep{UCIKDD}. In all cases, maximum likelihood joint
probability estimates were used. On sparse data, smoothing the probability
estimates is a good idea \citep{Good65}; we recommend multinomial
probability estimation with Dirichlet priors, a special case of which is
Laplace's law of succession; alternatively, for distributions following
the power law, such as word frequencies, Good-Turing smoothing is more
appropriate.

\subsection{Unsupervised Visualization}

We can illustrate the interaction quantities we discussed with
\emph{information graphs}, introduced in \citep{JakulinMAG} as interaction
diagrams.\footnote{Interaction diagram is a term frequently used for other
purposes, such as UML modelling, and this is why we renamed it.} They are
inspired by Venn diagrams, which we render as an ordinary graph, while the
surface area of each node identifies the amount of uncertainty. They can
also be seen as quantified factor graphs \citep{Kschischang01}.

White circles indicate the positive `information' of the model, the
entropy eliminated by the joint model. Gray circles indicate the two types
of negative `entropy', the initial uncertainty of the attributes and the
negative interactions indicating the redundancies. Redundancies can be
interpreted as overlapping of information, while information is
overlapping of entropy. The joint entropy of the attributes, or any subset
of them, is obtained by summing all the gray nodes and subtracting all the
white nodes linked to the relevant attributes.

\begin{figure}[htb]
\begin{center}
\includegraphics[width=9.5cm]{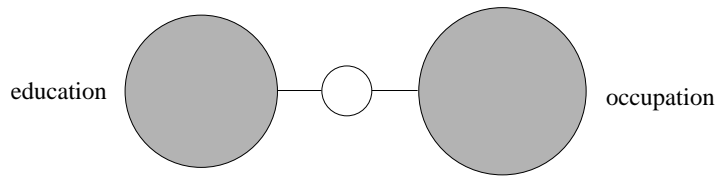}
\end{center}
\caption{Education and occupation do have something in common: the area of
the white circle indicates that the mutual information
$I(\text{education};\text{occupation})$ is non-zero. This is a 2-way
interaction, since two attributes are involved in it. The areas of the
gray circles quantify entropy of individual attributes:
$H(\text{education})$ and $H(\text{occupation})$. \label{fig:2}}
\end{figure}

We start with a simple example involving two attributes from the
`census/adult' data set, illustrated in Fig.~\ref{fig:2}. The instances
of the data set are a sample of adult population from a census database.
The occupation is slightly harder to predict a priori than the education
because occupation entropy is larger. Because the amount of mutual
information is fixed, the knowledge about the occupation will eliminate a
larger proportion of uncertainty about the level of education than vice
versa, but there is no reason for asserting directionality merely from the
data, especially as such predictive directionality could be mistaken for
causality.

\subsubsection{A negative interaction}

\begin{figure}[htb]
\begin{center}
\includegraphics[width=6.5cm]{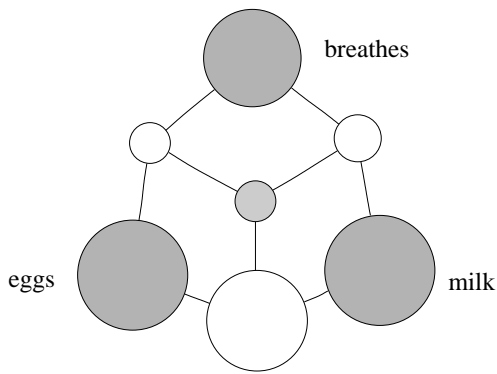}
\end{center}
\caption{An example of a 3-way negative interaction between the
properties `lays eggs?', `breathes?' and `has milk?' for different
animals. \label{fig:3-negative}}
\end{figure}

The relationship between three characteristics of animals in the `zoo'
database is rendered in Fig.~\ref{fig:3-negative}. All three attributes
are 2-interacting, but there is an overlap in the mutual information among
each pair, indicated by a negative interaction information. It is
illustrated as the gray circle, connected to the 2-way interactions, which
means that they have a shared quantity of information. It would be wrong
to subtract all the 2-way interactions from the sum of individual
entropies to estimate the complexity of the triplet, as we would
underestimate it. For that reason, the 3-way negative interaction acts as
a correcting factor.

This model is also applicable to supervised learning. If we were
interested if an animal breathes, but knowing whether it gives milk and
whether it lays eggs, we would obtain the residual uncertainty
$H(\text{breathes}|\text{eggs},\text{milk})$ by the following formula:
$$H(\text{breathes}) - \left(I(\text{breathes};\text{eggs}) +
I(\text{breathes};\text{milk}) +
I(\text{breathes};\text{eggs};\text{milk})\right).$$

This domain is better predictable than the one from Fig.~\ref{fig:2},
since the 2-way interactions are comparable in size to the prior attribute
entropies. It is quite easy to see that knowing whether whether an animal
lays eggs provides us pretty much all the evidence whether it has milk:
mammals do not lay eggs. Of course, such deterministic rules are not
common in natural domains.

Furthermore, the 2-way interactions between breathing and eggs and
between breathing and milk are very similar in magnitude to the 3-way
interaction, but opposite in sign, meaning that they cancel each other
out. Using the relationship between conditional mutual information and
interaction information from \eqref{eq:ig3}, we can conclude that:
\begin{align*}
    I(\text{breathes};\text{eggs}|\text{milk}) \approx & 0\\
    I(\text{breathes};\text{milk}|\text{eggs}) \approx & 0
\end{align*}
Therefore, if the 2-way interaction between such a pair is ignored, we
need no 3-way correcting factor. The relationship between these attributes
can be described with two Bayesian networks models, each assuming that a
certain 2-way interaction does not exist in the context of the remaining
attribute:
\begin{align*}
    \text{breathes} \leftarrow \text{milk} \rightarrow \text{eggs}\\
    \text{breathes} \leftarrow \text{eggs} \rightarrow \text{milk}
\end{align*}
If we were using the n\aive Bayesian classifier for predicting whether an
animal breathes, we might also find out that feature selection could
eliminate one of the attributes: Trying to decide whether an animal
breathes, and knowing that the animal lays eggs, most of the information
contributed by the fact that the animal doesn't have milk is redundant. Of
course, during classification we might have to classify an animal only
with the knowledge of whether it has milk, because the egg-laying
attribute value is missing: this problem is rarely a concern in feature
selection and feature weighting.

% Had we used feature construction, the milk and eggs attributes would be
% somewhat unexpectedly merged into their Cartesian product
% $\text{milk}\times\text{eggs}$, which is rather similar to using the
% following Bayesian network for classification:
% $$
%      \text{milk} \rightarrow \text{breathes} \leftarrow \text{eggs}.
% $$
% There is no indication here, however, that the reason for the dependence
% is simple correlation.

\subsubsection{A positive interaction}

\begin{figure}[htb]
\begin{center}
\includegraphics[width=8.5cm]{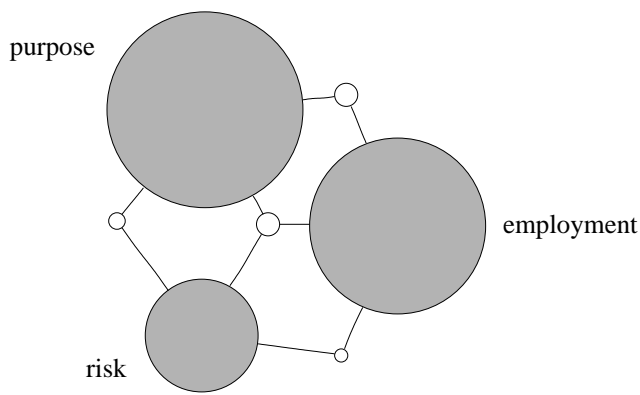}
\end{center}
\caption{An example of a 3-way positive interaction between the
customer's credit risk, his purpose for applying for a credit and his
employment status. \label{fig:3-positive}}
\end{figure}

Most real-life domains are difficult, meaning that it is hopeless trying
to predict the outcome deterministically. One such problem domain is a
potential customer's credit risk estimation. Still, we can do a good job
predicting the changes in risk for different attribute values. The `German
credit' domain describes credit risk for a number of customers.
Fig.~\ref{fig:3-positive} describes a relationship between the risk with a
customer and two of his characteristics. The mutual information between
any attribute pairs is low, indicating high uncertainty and weak
predictability. The interesting aspect is the positive 3-interaction,
which additionally reduces the entropy of the model. We emphasize the
positivity by painting the circle corresponding to the 3-way interaction
white, as this indicates information.

It is not hard to understand the significance of this synergy. On average,
unemployed applicants are riskier as customers than employed ones. Also,
applying for a credit to finance a business is riskier than applying for a
TV set purchase. But if we heard that an unemployed person is applying for
a credit to finance purchasing a new car, it would provide much more
information about risk than if an employed person had given the same
purpose. The corresponding reduction in credit risk uncertainty is the sum
of all three interactions connected to it, on the basis of employment, on
the basis of purpose, and on the basis of employment and purpose
simultaneously.

It is extremely important to note that the positive interaction coexists
with a mutual information between both attributes. If we removed one of
the attributes because it is correlated with the other one in a feature
selection procedure, we would also give up the positive interaction. In
fact, positively interacting attributes are often correlated.

\subsubsection{Interactions with zero interaction information}

\begin{figure}[htb]
\begin{center}
\includegraphics[width=8.5cm]{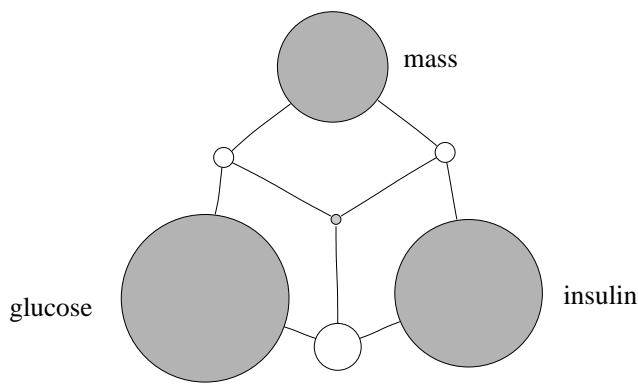}
\end{center}
\caption{An example of an approximately homogeneous association between
body mass, and insulin and glucose levels in the `pima' data set. All the
attributes are involved in 2-way interactions, yet the negative 3-way
interaction is very weak, indicating that all the 2-way interactions are
independent. An alternative explanation would be a mixture of a positive
and a negative interaction. \label{fig:3-homo}}
\end{figure}

The first explanation for a situation with zero 3-way interaction
information is that an attribute $C$ does not affect the relationship
between attributes $A$ and $B$, thus explaining the zero interaction
information $I(A;B|C) = I(A;B) \Rightarrow I(A;B;C) = 0$. A homogeneous
association among three attributes is described by all the attributes
2-interacting, but not 3-interacting. This would mean that their
relationship is fully described by a loopy set of 2-way marginal
associations. Although one could imagine that Fig.~\ref{fig:3-homo}
describes such a homogeneous association, there is another possibility.

Imagine a situation which is a mixture of a positive and a negative
interaction. Three attributes $A,B,C$ take values from $\{0,1,2\}$. The
permissible events are $\{e_1:A=B+C\pmod 2, \, e_2:A=B=C=2\}$. The event
$e_1$ is the familiar XOR problem, denoting a positive interaction. The
event $e_2$ is an example of perfectly correlated attributes, an example
of a negative interaction. In an appropriate probabilistic mixture, for
example $\text{Pr}\{e_1\}\approx 0.773, \, \text{Pr}\{e_2\}\approx 0.227$,
the interaction information $I(A;B;C)$ approaches zero. Namely, $I(A;B;C)$
is the average interaction information across all the possible
combinations of values of $A,B$ and $C$. The distinctly positive
interaction for the event $e_1$ is cancelled out, on average, with the
distinctly negative interaction for the event $e_2$. The benefit of
joining the three attributes and solving the XOR problem exactly matches
the loss caused by duplicating the dependence between the three
attributes.

Hence, 3-way interaction information should not be seen as a full
description of the 3-way interaction but as the interaction information
averaged over the attribute values, even if we consider interaction
information of lower and higher orders. These problems are not specific
only to situations with zero interaction information, but in general. If a
single attribute contains information about complex events, much
information is blended together, which should rather be kept apart. Not to
be misled by such mixtures, we may represent a many-valued attribute $A$
with a set of binary attributes, each corresponding to one of the values
of $A$. Alternatively, we may examine the value of interaction information
at particular attribute values. The visualization procedure may assist in
determining the interactions to be examined closely by including bounds or
confidence intervals for interaction information across all combinations
of attribute values; when the bounds are not tight, a mixture can be
suspected.

\subsubsection{Interaction patterns in data}

If the number of attributes under investigation is increased, the
combinatorial complexity of interaction information may quickly get out of
control. Fortunately, interaction information is often low for most
combinations of unrelated attributes. We have also observed that the
average interaction information of a certain order is decreasing with the
order in a set of attributes. A simple approach is to identify $N$
interactions with maximum interaction magnitude among the $n$. For
performance and reliability, we also limit the maximum interaction order
to $k$, meaning that we only investigate $l$-way interactions, $2 \leq l
\leq k \leq n$. Namely, it is difficult to reliably estimate joint
probability distributions of high order. The estimate of $P(X)$ is usually
more robust than the estimate of $P(X,Y,Z,W)$ given the same number of
instances.

\paragraph{Mediation and moderation}

\begin{figure}[htb]
\begin{center}
\includegraphics[width=10cm]{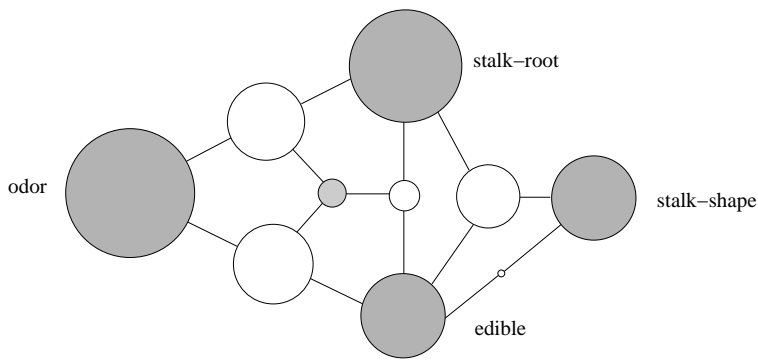}
\vspace{-5mm}
\end{center}
\caption{A selection of several important interactions in the `mushroom'
domain. \label{fig:lilmush}}
\end{figure}

A larger scale information graph with a selection of interactions in the
`mushroom' domain is illustrated in Fig.~\ref{fig:lilmush}. Because
edibility is the attribute of interest (the label), we center our
attention on it, and display a few other attributes associated with it.
The informativeness of the stalk shape attribute towards mushroom's
edibility is very weak, but this attribute has a massive synergistic
effect if accompanied with the stalk root shape attribute. We can describe
the situation with the term \emph{moderation}~\citep{BaronKenny86}: stalk
shape `moderates' the effect of stalk root shape on edibility. Stalk shape
is hence a moderator variable. It is easy to see that such a situation is
problematic for feature selection: if our objective was to predict
edibility, a myopic feature selection algorithm would eliminate the stalk
shape attribute, before we could take advantage of it in company of stalk
root shape attribute. Because the magnitude of the mutual information
between edibility and stalk root shape is similar in magnitude to the
negative interaction among all three, we can conclude that there is a
conditional independence between edibility of a mushroom and its stalk
root shape given the mushroom's odor. A useful term for such a situation
is \emph{mediation}~\citep{BaronKenny86}: odor `mediates' the effect of
stalk root shape on edibility.

The 4-way interaction information involving all four attributes was
omitted from the graph, but it is distinctly negative. This can be
understood by looking at the information gained about edibility from the
other attributes and their interactions with the actual entropy of
edibility: we cannot explain 120\% of entropy, unless we are counting the
evidence twice. The negativity of the 4-way interaction indicates that a
certain amount of information provided by the stalk shape, stalk root
shape and their interaction is also provided by the odor attribute.

\paragraph{Synonymy and polysemy}

\begin{figure}[htb]
\begin{center}
\includegraphics[width=8.5cm]{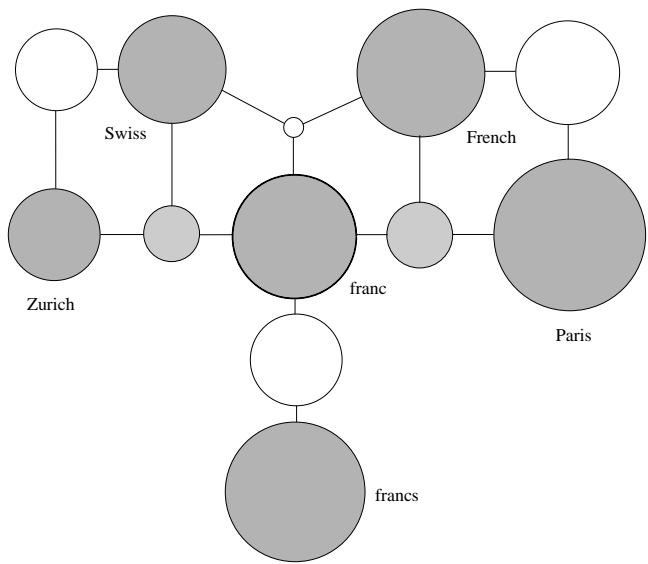}
\vspace{-4mm}
\end{center} \caption{A selection of interactions involving
the keyword `franc' in news reports shows that interaction analysis can
help identify useful contexts, synonyms and polysemy in information
retrieval. \label{fig:reuters}}
\end{figure}

In complex data sets, such as the ones for information retrieval, the
number of attributes may be measured in tens of thousands. Interaction
analysis must hence stem from a particular reference point. For example,
let us focus on the keyword `franc', the currency, in the `Reuters' data
set. This keyword is not a label, but merely a determiner of context. We
investigate the words that co-appear with it in news reports, and identify
a few that are involved in 2-way interactions with it. Among these, we may
identify those of the 3-way interactions with high normed interaction
magnitude. The result of this analysis is rendered in
Fig.~\ref{fig:reuters}. We can observe the positive interaction among
`Swiss', `French' and `franc' which indicates that `franc' is polysemous.
There are two contexts in which the word `franc' appears, but these two
contexts do not mix, and this causes the interaction to be positive. The
strong 2-way interaction between `franc' and `francs' indicates a
likelihood of synonymy: the two words are frequently both present or both
absent, and the same is true of pairs `French'-`Paris' and
`Swiss'-`Zurich'. Looking at the mutual information (which is not
illustrated), the two negative interactions are in fact near conditional
independencies, where `Zurich' and `franc' are conditionally independent
given `Swiss', while `French' and `franc' are conditionally independent
given `Paris'. Hence, the two keywords that are best suited to distinguish
the contexts of the kinds of `franc' are `Swiss' and `Paris'. These tree
are positively interacting, too.

\subsection{Supervised Visualization}

The objective of supervised learning is acquiring information about a
particular label attribute from the other attributes in the domain. In
such circumstances, we are interested only in those relationships between
attributes that involve the label. Figuratively, we place ourselves into
the label and view the other attributes from this perspective. It enables
us to simplify the earlier diagrams considerably, which, in turn,
facilitates application of interaction analysis methodology to exploratory
data analysis.

There are several types of inter-attribute relationships which can be of
interest. \emph{Interaction graphs}~\citep{Jakulin03b} disclose 2-way and
3-way interactions involving the label in a domain. \emph{Interaction
dendrograms}~\citep{Jakulin03a} are a compact summary of proximity
between attributes with respect to the similarity (or synergy) of the
information they provide about the label. \emph{Conditional interaction
graphs} attempt to illustrate the magnitude of unwanted dependencies
which affect the performance in learning algorithms that make the
conditional independence assumption.

\subsubsection{Interaction dendrograms}

In initial phases of exploratory data analysis, we might not be interested
in detailed relationships between attributes, but merely wish to discover
groups of mutually interacting attributes. In supervised learning, we are
not investigating the relationships between attributes themselves (where
mutual information would have been the metric of interest), but rather the
relationships between the mutual information of either attribute with the
label. In other words, we would like to know whether two attributes
provide similar information about the label, or whether there is synergy
between attributes' information about the label.

To perform any kind of similarity-based analysis, we should define a
similarity or a dissimilarity measure between attributes. With respect to
the amount of interaction, interacting attributes should appear close to
one another, and non-interacting attributes far from one another. One of
the most frequently used similarity measures for clustering is
\emph{Jaccard's coefficient} \citep{Jaccard08}. For two sets, $\cal A$ and
$\cal B$, the Jaccard's coefficient (along with several other similarity
measures) can be expressed through set cardinality \citep{StatNLP99}:
\begin{equation}
J({\cal A},{\cal B}) \triangleq \frac{\left| {\cal A} \cap {\cal
B} \right|}{\left| {\cal A} \cup {\cal B} \right|}
\end{equation}
If we understand interaction information as a way of measuring the
cardinality of the intersection in Fig.~\ref{fig:venn2} and in
Section.~\ref{sect:set-entropy}, where mutual information corresponds to
the intersection, and joint entropy as the union, we can define the
\emph{normed mutual information} between attributes $A$ and $B$:
\begin{equation}
\|I(A,B)\| \triangleq \frac{I(A;B)}{H(A,B)}
\end{equation}
M\`{a}ntaras' distance $d_M$ \citep{Mantaras91}, which has been shown to
be a useful heuristic for feature selection, less sensitive to the
attribute alphabet size, is closely related to normed mutual information:
$d_M(A,B)=1-\|I(A,B)\|$. Dividing by the joint entropy helps us reduce the
effect of the number of attribute values, hence facilitating comparisons
of mutual information between different attributes. Rajski's distance
\citep{Rajski61} is identical to $d_M$, but predates it, while normed
mutual information is identical to interdependence redundancy
\citep{Wong75}.

To present the attribute information similarities to a human analyst, we
may tabulate it in a dissimilarity matrix, where color may code the
interaction information magnitude and interaction type. Such matrices may
become unwieldy with a large number of attributes. To summarize them, we
employ the techniques of clustering or multi-dimensional scaling. For
visualizing higher-order interactions in an analogous way, we can
introduce a further attribute $C$ either as context, using normed
conditional mutual information $\|I(A,B|C)\| \triangleq
I(A;B|C)/H(A,B|C)$, or as another attribute in interaction information by
using normed interaction magnitude:
\begin{equation}
\|I(A,B,C)\|_+ \triangleq \frac{\left|I(A;B;C)\right|}{H(A,B,C)},
\end{equation}
where the interaction magnitude $\left|I(A;B;C)\right|$ is the absolute
value of interaction information. $C$ has to be fixed and usually
corresponds to the label, while $A$ and $B$ are variables that iterate
across all combinations of remaining attributes.

\begin{figure}[htb]
\begin{center}
\includegraphics[width=9.5cm]{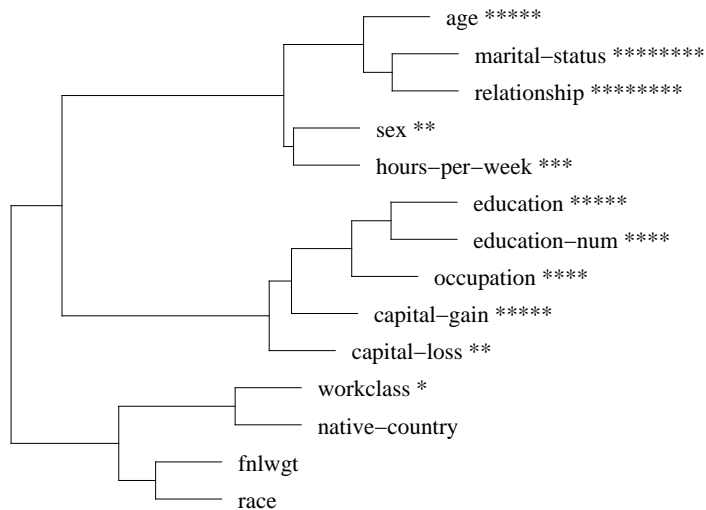}
\end{center}
\caption{An interaction dendrogram illustrates which attributes interact,
positively or negatively, with the label in the `census/adult' data set.
The label indicates the individual's salary. The number of asterisks
indicates the amount of mutual information between an attribute and the
label. \label{fig:dendrogram}}
\end{figure}

In the example in Fig.~\ref{fig:dendrogram}, we used Ward's method for
agglomerative hierarchical clustering for summarizing the attribute
similarity matrix of normed interaction magnitude $\|I(A,B,C)\|_+$ for
every pair of attributes $A,B$ and the fixed label $C$. Since the
clustering algorithm worked with dissimilarities rather than with
similarities, we took the reciprocal value of the normed interaction
magnitude, and set an upper limit of dissimilarity (e.g., $K=1000$) to
prevent independent attributes from disproportionately affecting the
graphical representation.

The resulting interaction dendrogram is one approach to variable
clustering, where the proximity is based on the redundancy or synergy of
the attributes' information about the label. We can observe that there are
two distinct clusters of attributes. One cluster contains attributes
related to the lifestyle of the person: age, family, working hours, sex.
The second cluster contains attributes related to the occupation and
education of the person. The third cluster is not compact, and contains
the information about the native country, race and work class, all
relatively uninformative about the label.

Normed interaction magnitude helps identify the groups of attributes that
should be investigated more closely. We can use color to convey the type
of the interaction. For example, we color zero interactions green,
positive interactions red and negative interactions blue, mixing all three
color components depending on the normed interaction information. Blue
clusters indicate on average negatively interacting groups, and red
clusters indicate positively interacting groups of attributes.

Interaction dendrograms may be useful for feature selection. The fnlwgt
and race attributes in Fig.~\ref{fig:dendrogram} do not participate in any
positive interactions, and are uninformative by themselves: they are
natural candidates for elimination during feature selection. On the other
hand, each cluster of negatively interacting attributes has a considerable
amount of redundance. For example, older people tend to be married, and
highly educated people have spent many years in school. In aggressive
feature selection, we could hence simply pick the individually best
attributes from each cluster. In our example, these would be marital
status and education.

\subsubsection{Interaction graphs}

Many interesting relationships are not visible in detail in the
dendrogram. To drill deeper into the relationships among a group of
attributes, we can apply interaction graphs. There, individual attributes
are represented as graph nodes and a selection of the 3-way interactions
as edges. To limit the complexity arising from a combinatorial explosion
of the number of interactions in a domain with many attributes, we
restrict the number of interactions illustrated only to those with the
largest magnitude, and to those that involve the label.

\begin{figure}[htb]
\begin{center}
\includegraphics[width=9.5cm]{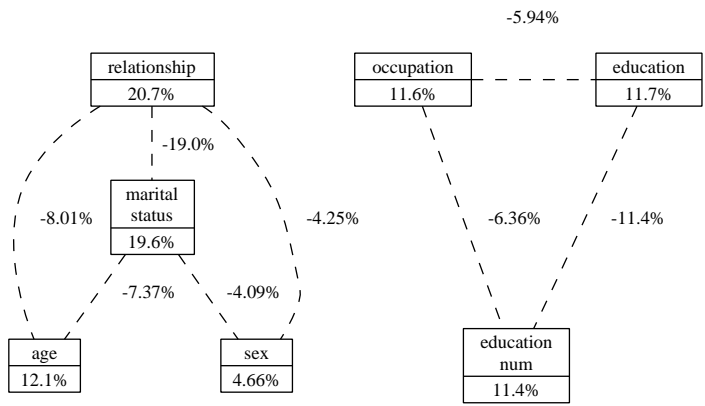}
\vspace{-7mm}
\end{center}
\caption{An interaction graph containing eight of the 3-way interactions
with the largest interaction magnitude in the `adult/census' domain. For
all pairs, the interaction information is negative.
\label{fig:adult-graph}}
\end{figure}

Nodes and edges of an interaction graph are labelled numerically. The
percentage in the node expresses the amount of label's uncertainty
eliminated by the the node's attribute, the relative mutual information.
For example, in Fig.~\ref{fig:adult-graph}, the most informative attribute
is relationship (describing the role of the individual in his family), and
the mutual information between the label and relationship amounts to
$20.7\%$ of salary's entropy.

The dashed edges indicate negative interactions that involve the two
connected attributes and the label. They are too labelled with relative
interaction information, for example, the negative interaction between
relationship, marital status and the label comprises $19\%$ of the label's
entropy. If we wanted to know how much information we gained about the
salary from these two attributes, we would sum up the mutual information
for both 2-way interactions and the 3-way interaction information:
$20.7+19.6-19 = 21.3\%$ of entropy was eliminated using both attributes.
Once we knew the relationship of a person, the marital status further
eliminated only $0.6\%$ of the salary's entropy.

The interaction graph is an approximation to the true relationships
between attributes, as only a part of 3-way interactions are drawn,
without regard to interactions of higher order. The number of interactions
illustrated was determined merely on the basis of graph clarity: if the
graph became cluttered, we reduced the number of interactions shown.
Therefore, we should be careful when generalizing the above entropy
computations to more than three attributes. We have observed that negative
interactions, viewed as relations, tend to be transitive, but not positive
interactions. Namely, $I(A;B;C)$ can only be negative if some 2-way mutual
information, e.g., $I(A;B)$ is large in comparison to $I(A;B|C)$. This
large mutual information will also be large in several other 3-way
interactions that include attributes $A$ and $B$. On the other hand, a
positive 3-way interaction information is determined by the specifics of
all three attributes, and no other 3-way interaction involves these three
attributes.

\begin{figure}[htb]
\begin{center}
\includegraphics[width=10.5cm]{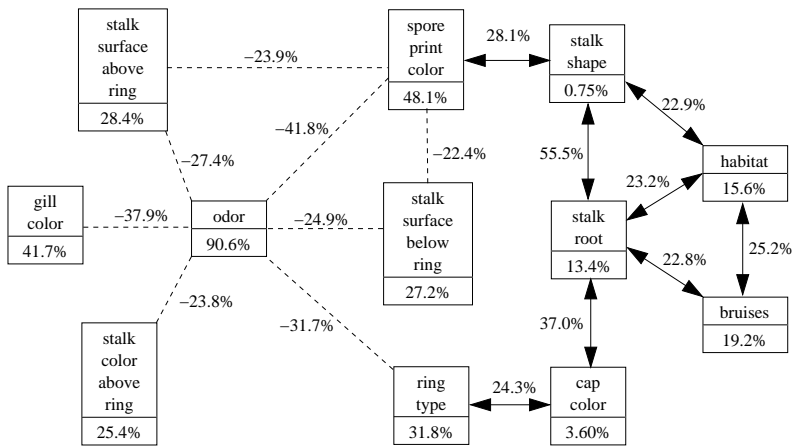}
\vspace{-5mm}
\end{center}
\caption{An interaction graph containing eight of the positive and eight
of the negative 3-way interactions with the largest interaction magnitude
in the `mushroom' domain. The positive interactions are indicated by
solid arrows. \label{fig:mushroom-graph}}
\end{figure}

As an example of a more complex interaction graph we illustrate the
familiar `mushroom' domain in Fig.~\ref{fig:mushroom-graph}. As an
example, let us consider the positive interaction between stalk and stalk
root shape. Individually, stalk root shape eliminates $13.4\%$, while
stalk shape only $0.75\%$ of edibility entropy. If we exploit the
synergy, we gain additional $55.5\%$ of entropy. Together, these two
attributes eliminate almost $70\%$ of our uncertainty about a mushroom's
edibility.

\subsubsection{Conditional interaction graphs}

The visualization methods described in previous sections are unsupervised
in the sense that they describe 3-way interactions outside any context,
for example $I(A;B;C)$. They were merely customized for the properties of
analysis specific to supervised learning. We now focus on conditional
interactions that affect model selection in supervised learning, such as
$I(A;B|Y)$ and $I(A;B;C|Y)$ where $Y$ is the label. These are useful for
verifying the grounds for taking the conditional independence assumption
in the n\aive Bayesian classifier. Such assumption may be problematic if
there are informative conditional interactions between attributes with
respect to the label.

\begin{figure}
\begin{center}
\includegraphics[width=13.5cm]{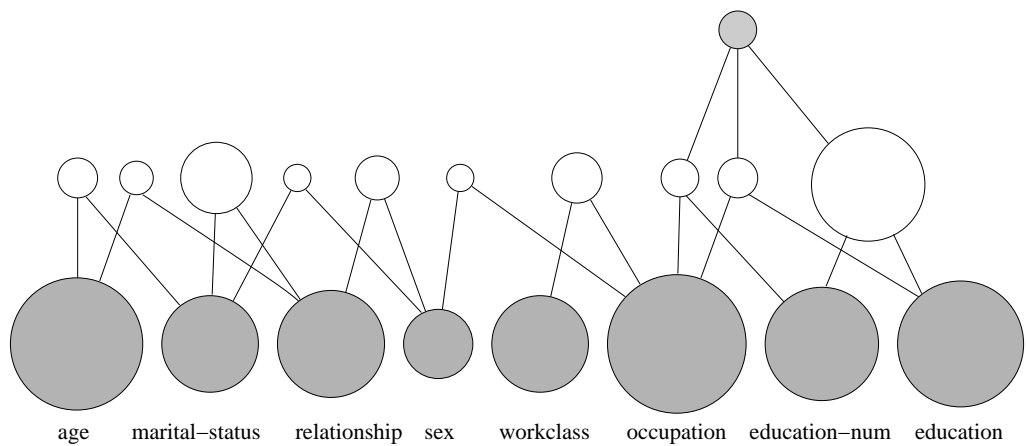}
\end{center}
\caption{An informative conditional interaction graph illustrating the
informative conditional interactions between the attributes with respect
to the salary label in the `adult/census' domain. \label{fig:cond-adult}}
\end{figure}

In Fig.~\ref{fig:cond-adult} we have illustrated the informative
conditional interactions with large magnitude in the `adult/census' data
set, with respect to the label -- the salary attribute. Learning with the
conditional independence assumption would imply that these interactions
are ignored. The negative 3-way conditional interaction with large
magnitude involving education, years of education and occupation (in the
context of the label) offers a possibility for simplifying the domain.
Other attributes from the domain were left out from the chart, as only the
race and the native country attribute were conditionally interacting.

\section{Handling Interactions in Supervised Learning}
\label{sec:learning-proc}

The machine learning community has long been aware of interactions, and
many methods have been developed to deal with them. There are two problems
that may arise from incorrect treatment of interactions: \emph{myopia} is
the consequence of assuming that interactions do not exist, even if they
do exist; \emph{fragmentation} is the consequence of acting as if
interactions existed, when they are not significant. We will briefly
survey several popular learning techniques in the light of the role they
have with respect to interactions, and provide guidelines.

\subsection{Myopia}

Greedy feature selection and split selection heuristics are often based on
various quantifications of 2-way interactions between the label $Y$ and an
attribute $A$. The frequently used information gain heuristic in decision
tree learning is a simple example of how interaction magnitude has been
used for evaluating attribute importance. With more than a single
attribute, information gain is no longer a reliable measure. First, with
positive interactions, such as the exclusive or problem, information gain
may underestimate the actual importance of attributes, since
$I(A,B;Y)>I(A;Y)+I(B;Y)$. Second, in negative interactions, information
gain will overestimate the importance of attributes, because some of the
information is duplicated, as can be seen from $I(A,B;D)<I(A;D)+I(B;D)$.

These problems with positive interactions are known as \emph{myopia}
\citep{kononenko97overcoming}. Myopic feature selection evaluates an
attribute's importance independently of other attributes, and it is unable
to appreciate their synergistic effect. The inability of myopic feature
selection algorithms to appreciate interactions can be remedied with
algorithms such as Relief~\citep[e.g.][]{Kira,RobnikMLJ}, which increase
the estimated quality of positively interacting attributes, and reduce the
estimated worth of negatively interacting attributes.

Ignoring negative interactions may cause several problems in machine
learning and statistics. We may end up with attributes providing the same
information multiple times, hence biasing the predictions. For example,
assume that the attribute $A$ is a predictor of the outcome $Y_0$, whereas
the attribute $B$ predicts the outcome $Y_1$. If we duplicate $A$ into
another attribute $A'$ but retain a sole copy of $B$, n\aive Bayesian
classifier trained on $\{A,A',B\}$ will be biased towards the outcome
$Y_0$. Hence, negative interactions offer opportunity for eliminating
redundant attributes, even if these attributes are informative on their
own. An attribute $A'$ would then be a conditionally irrelevant source of
information about the label $Y$ given the attribute $A$ when $I(A';Y|A) =
0$, assuming that there are no other attributes positively interacting
with the disposed attribute \citep{Koller96}. Indirectly, we could
minimize the mutual information among the selected attributes, via
eliminating $A'$ if $I(A;A')$ is large \citep{Hall00}. Finally, attribute
weighting, either explicit (by assigning weights to attributes) or
implicit (such as fitting logistic regression models or support vector
machines), helps remedy some examples of negative interactions. Not all
examples of negative interactions are problematic, however, since
conditional independence between two attributes given the label may result
in a negative interaction information among all three.

Feature selection algorithms are not the only algorithms in machine
learning that suffer from myopia. Most supervised discretization
algorithms \citep[e.g.][]{FayyadIrani93} are local and discretize one
attribute at a time, determining the number of intervals with respect to
the ability to predict the label. Such algorithms may underestimate the
number of intervals for positively interacting attributes
\citep{Nguyen98,Bay01}. For example, in a domain with two continuous
attributes $X$ and $Y$, labelled with class $Z_1$ when $X>0,\, Y>0$ or
$X<0,\, Y<0$, and with class $Z_0$ when $X>0,\, Y<0$ or $X<0,\, Y>0$ (the
continuous version of the binary exclusive or problem), all univariate
splits are uninformative. On the other hand, for negatively interacting
attributes, the total number of intervals may be larger than necessary,
causing fragmentation of the data. Hence, in case of positive and negative
interactions, multivariate or global discretization algorithms may be
preferred.

\subsection{Fragmentation}

To both take advantage of synergies and prevent redundancies, we may use a
different set of more powerful methods. We may assume dependencies between
attributes by employing dependence modelling
\citep{kononenko91,Friedman97}, create new attributes with structured
induction methods \citep{Shapiro87,Pazzani96,Zupan99}, or create new
classes via class decomposition \citep{Vilalta03}. The most frequently
used methods, however, are the classification tree and rule induction
algorithms. In fact, classification trees were originally designed also
for detection of interactions among attributes in data: one of the first
classification tree induction systems was named Automatic Interaction
Detector (AID) \citep{AID1}.

Classification trees are an incremental approach to modelling the joint
probability distribution $P(Y|A,B,C)$. The information gain split
selection heuristic \citep[e.g.][]{Quinlan86} seeks the attribute $A$ with
the highest mutual information with the label $Y$: $A =
\arg\max_X{I(Y;X)}$. In the second step, we pursue the attribute $B$,
which will maximize the mutual information with the label $Y$, but in the
context of the attribute $A$ selected earlier: $B = \arg\max_X{I(Y;X|A)}$.

In case of negative interactions between $A$ and $B$, the classification
tree learning method will correctly reduce $B$'s usefulness in the context
of $A$, because $I(B;Y|A) < I(B;Y)$. If $A$ and $B$ interact positively,
$B$ and $Y$ will have a larger amount of mutual information in the context
of $A$ than otherwise, $I(B;Y|A) > I(B;Y)$. Classification trees enable
proper treatment of positive interactions between the currently evaluated
attribute and the other attributes already in the context. However, if the
other positively interacting attribute has not been included in the tree
already, then this positive 2-way interaction may be overlooked. To assure
that positive interactions are not omitted, we may construct the
classification tree with look-ahead \citep{Norton89,Ragavan93}, or we may
seek interactions directly \citep{Perez97}.

The classification tree learning approach does handle interactions, but it
is not able to take all the advantage of mutually and conditionally
independent attributes. Assuming dependence increases the complexity of
the model because the dimensionality of the probability distributions
estimated from the data is increased. A consequence of this is known as
\emph{fragmentation} \citep{Vilalta97}, because the available mutual
information between an attribute $B$ and the label $Y$ is not assessed on
all the data, but merely on fragments of it. Fragmenting is harmful if the
context $A$ is independent of the interaction between $B$ and $Y$. For
example, if $I(B;Y) = I(B;Y|A)$, the information provided by $B$ about $Y$
should be gathered from all the instances, and not separately in each
subgroup of instances with a particular value of the attribute $A$. This
is especially important when the training data is scarce. Although we used
classification trees as an example of a model that may induce
fragmentation, other methods too are subject to fragmentation by assuming
dependence unnecessarily.

Three approaches may be used to remedy fragmentation. One approach is
based on ensembles: aggregations of simpler trees, each specializing in a
specific interaction. For example, random forests \citep{RandomForest}
aggregate the votes from a large number of small trees, where each tree
can be imagined to be focusing on a single interaction. One can use hybrid
methods that employ both classification trees and linear models that
assume conditional independence, such as the n\aive Bayesian classifier
\citep{Kononenko88,Kohavi96} or logistic regression
\citep{AbuHanna03,Landwehr03}. Finally, feature construction algorithms
may be employed in the context of classification tree induction
\citep[e.g.][]{Pagallo90,SetionoLiu}.

%%%%%%%%%%%%%%%%%%%%%%%%%%%%%%%%%%%%%%%%%%%%%%%%%%%%%%%%%%%%%%%%%%%%%%%%%%%%%%%%%%%%

\section{Summary and Discussion}

We have defined an interaction in a set of attributes to be the loss we
incur by approximating the joint probability distribution of the
attributes by only using marginal probability distributions. This notion
is captured by McGill's interaction information, a special case of which
is mutual information. Interaction information has been independently
rediscovered a number of times in a variety of fields, including machine
learning, computational neuroscience, psychology and information and game
theory. Interaction information can also be understood as a nonlinear
generalization of correlation between any number of attributes. We can
distinguish positive and negative interactions, depending on the sign of
interaction information. Positive interactions indicate phenomena such as
moderation, where one attribute affects the relationship of other
attributes. On the other hand, negative interactions may suggest
mediation, where one or more of the attributes in part convey the
information already provided by other attributes.

The goal of interaction analysis is quantification of interactions and
presentation of the interaction structure in the data in a comprehensible
form to a human analyst. To this aim we have proposed a number of novel
visualization methods that attempt to present interactions present in
data, given some quantification of interactions. The interaction
dendrogram is a compact summary of the attribute structure, identifying
clusters of negatively interacting attributes, and connecting pairs of
positively interacting attributes. The interaction graph identifies the
most important interactions in detail. The information graph is a
substitute for the Venn diagram and it illustrates the detailed structure
of dependence in a small set of attributes. We show on numerous examples
that the above quantifications usually confirm the intuitions.

There are two pitfalls that a good supervised learning procedure should
avoid. The problem of myopia arises when a learning algorithm assumes that
interactions do not exist, but they do. This results in the classifier
\emph{bias}. In myopic learning, negative interactions result in redundant
models, while synergies between attributes are not taken advantage of. On
the other hand, fragmentation is a consequence of assuming interactions
when they do not exist or are not important enough. Fragmentation induces
a learning procedure to gather statistics from less data than it could,
causing unreliable estimates of evidence and the classifier
\emph{variance}. Fragmentation is also an issue for detecting
interactions. To detect an interaction, we need to estimate the joint
probability distribution of several variables. For this, a considerable
amount of data is needed. For example, if one tries to do 3-way
interaction analysis with only 100 instances, there will be a lot of noisy
positive interactions, but few of them are significant. To solve this
problem, the methods of statistical inference may be employed, for example
hypothesis testing.

We are currently researching learning algorithms that are not sidetracked
by or blind of interactions. Namely, pursuing interactions in data is
complementary to pursuit of independence, but from the opposite direction.
Starting with a simple model with no dependencies, we gradually build a
complex one by successively introducing important interactions. One
important problem in this context is the combinatorial explosion of the
attribute combinations. However, one can employ heuristics, for example,
higher-order interactions are unlikely in the absence of lower-order
interactions among the same attributes. We are also developing methodology
for investigating continuous attributes. Finally, in this text, we have
made several assumptions, which are not always justified: attribute values
are not always mutually independent, the instances are not always
permutable. Relaxing these assumption is also an area for future research.

\acks{The authors wish to thank A. J. Bell and T. P. Minka for helpful
comments and discussions, to S. Ravett Brown for providing the original
McGill's paper, to J. Dob\v{s}a for drawing our attention to the issue of
synonymy and polysemy, and to V. Batagelj for pointing us to Rajski's
work.}

\vskip 0.2in
\bibliography{../bib/interactions}

\end{document}